\documentclass[11pt]{article}

\usepackage[final]{acl}

\usepackage{times}
\usepackage{latexsym}

\usepackage[T1]{fontenc}
\usepackage[utf8]{inputenc}

\usepackage{microtype}

\usepackage{inconsolata}

\usepackage{graphicx}

\usepackage{subcaption}
\usepackage{amsmath}
\usepackage{booktabs}
\usepackage{tabularx}
\usepackage{multirow}
\usepackage[table]{xcolor}
\definecolor{headerblue}{RGB}{30, 55, 100}
\definecolor{comprow}{RGB}{240, 240, 245}
\definecolor{ourgreen}{RGB}{231, 241, 232}
\definecolor{groupgray}{gray}{0.95}
\newcolumntype{Y}{>{\centering\arraybackslash}X}
\usepackage{array}
\usepackage{algorithm}
\usepackage{algpseudocode}

\title{ROT: Rotating Hidden States towards Contextual Vectors for Hallucination Mitigation in LVLMs}

\author{
  \textbf{Yijing Du} \quad
  \textbf{Xiangcheng Zhan} \quad
  \textbf{Shuo Yang}\thanks{Corresponding author} \\
  Harbin Institute of Technology, Shenzhen \\
  \small{
    \texttt{2023311202@stu.hit.edu.cn} \quad
    \texttt{25b951051@stu.hit.edu.cn} \quad
    \texttt{shuoyang@hit.edu.cn}
  }
}

\begin{document}
\maketitle
\begin{abstract}
Large Vision-Language Models (LVLMs) frequently suffer from object hallucination. Existing training-free interventions primarily manipulate attention weights, which indirectly affect the deep semantics reaching the final predictive layers. In this work, we shift our focus to the hidden state vectors extracted after self-attention and residual addition. Empirical analysis reveals that hallucinated tokens do not simply over-rely on linguistic priors; instead, they exhibit an anomalous contextual deviation, showing significantly lower similarities to both textual and visual contexts in intermediate layers. Motivated by this, we propose ROT, a layer-specific, training-free framework. ROT dynamically detects semantic deviation in the middle layers and applies a norm-preserving rotation to steer the hidden states back toward the local multimodal context plane spanned by the contexts. For subsequent layers, a representational smoothing mechanism is introduced to stabilize the calibrated trajectory. Extensive experiments on multiple benchmarks demonstrate that ROT consistently reduces hallucinations across various model architectures and scales, offering an efficient, geometry-driven solution for grounded generation.
\end{abstract}

\section{Introduction}
Large Vision-Language Models (LVLMs) have demonstrated exceptional capabilities in integrating visual perception with linguistic reasoning. However, these models are persistently plagued by object hallucination, generating descriptions of objects or attributes that are inconsistent with the visual input \cite{li2023evaluating, zhou2023analyzing}. This fundamental flaw poses severe reliability risks for real-world applications. While extensive efforts have been made to mitigate hallucinations via instruction tuning, such training-based paradigms are often computationally prohibitive. Consequently, training-free inference-time interventions have emerged as an efficient alternative to dynamically calibrate model outputs.

Recent inference-time interventions have made notable strides in this direction. For instance, some approaches adjust visual information propagation in the middle layers \cite{jiang2025devils}, selectively drop visual tokens based on epistemic uncertainty \cite{fang2024enhancing}, or leverage gradient-aware saliency to filter ungrounded tokens \cite{zhang2026hallucination}. Despite their effectiveness, these methods primarily manipulate attention weights or passively filter input contexts. We argue that solely adjusting attention weights does not directly determine the actual semantic content reaching the final predictive layers. To effectively rectify hallucinations, we shift our focus to the hidden state vectors after self-attention and before the feed-forward network.

To understand the dynamics of these hidden states during hallucination, we extract the normalized attention matrix at a given layer $k$ to aggregate the hidden states of the previous layer, yielding a textual context vector $C_T$ and a visual context vector $C_V$. Under a simple language-prior explanation, one might expect hallucinated tokens to exhibit stronger alignment with the textual context $C_T$. However, our empirical analysis shows that the hidden states of hallucinated tokens $h_k$ exhibit lower cosine similarities with both $C_T$ and $C_V$. This observation resonates with recent findings that hallucinations often begin where contextual memory or saliency drops \cite{zhang2026hallucination}. These results suggest that, for the analyzed object-hallucination cases, over-reliance on textual priors alone is insufficient to characterize the observed hidden-state behavior; instead, hallucinated representations exhibit a simultaneous contextual deviation from both modality-specific anchors.

\begin{figure*}[t]
    \centering
    \includegraphics[width=\textwidth]{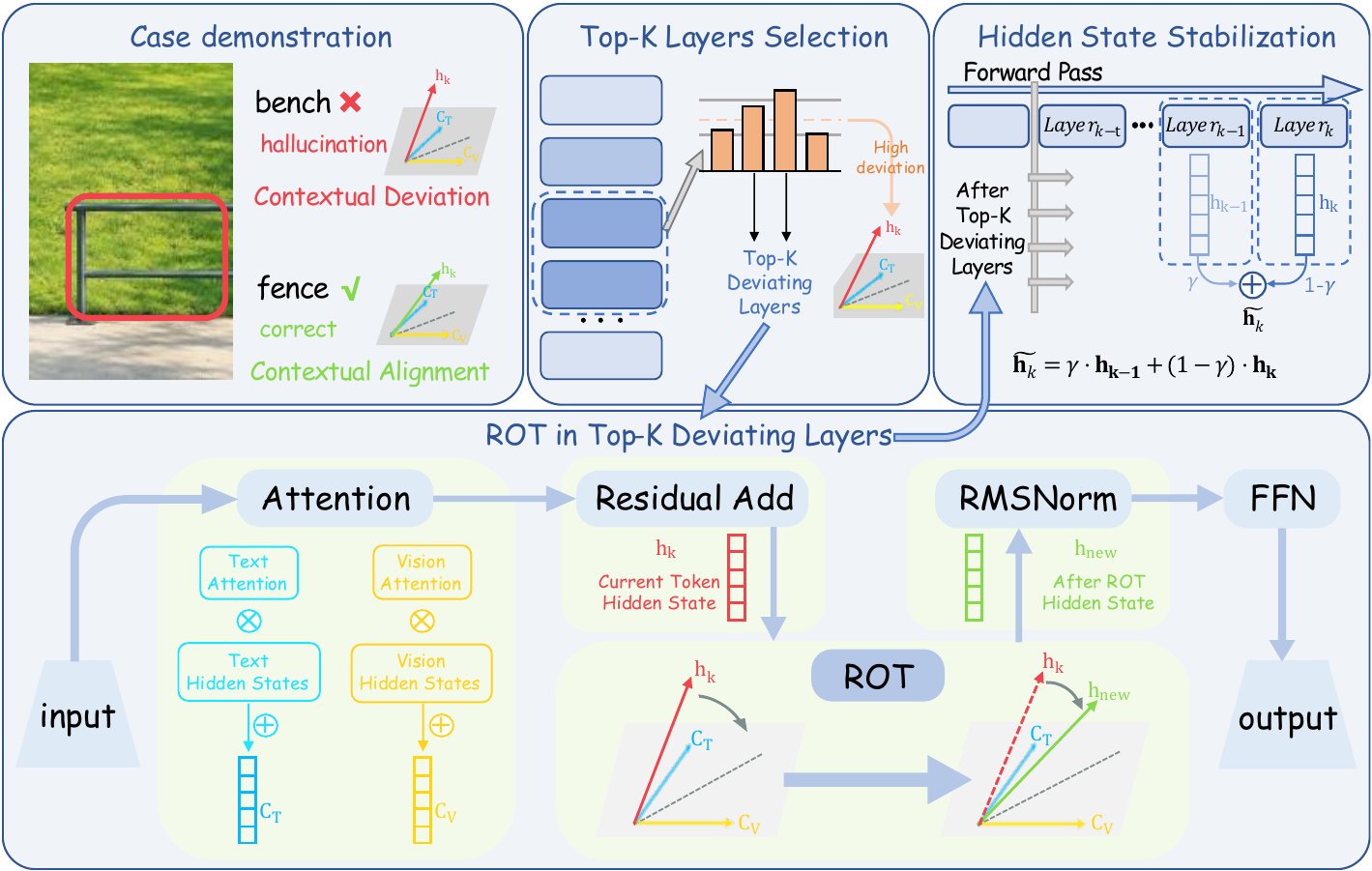}
    \caption{The overall framework of the proposed ROT method. The process consists of two stages: (1) Dynamic Hallucination Detection in the critical middle layers, which identifies contextual deviation based on the similarities between the hidden state and multimodal contexts; and (2) Layer-Specific Hidden State Mitigation, which applies a norm-preserving rotation in the intermediate layers to steer the deviating representation back toward the grounded factual plane, followed by a representational smoothing mechanism in subsequent layers to maintain stability during propagation.}
    \label{fig:main_image}
\end{figure*}

Recognizing this behavior, we propose \textbf{ROT}, \textbf{a training-free framework featuring layer-specific hidden state intervention}. As shown in Figure \ref{fig:main_image}, we formulate a deviation indicator $E$, which is derived from the complement of the cosine similarities between the current hidden state $h_k$ and the two context vectors $C_T$ and $C_V$. Our layer-wise analysis reveals that the value of $E$ is predominantly high in the middle transformer layers, indicating a high propensity for hidden states to experience severe semantic deviation in these intermediate stages. Consequently, we specifically target the middle layers for hallucination detection. When $E$ exceeds a predefined threshold, it signifies an erroneous semantic deviation, thereby triggering the mitigation mechanism.

For mitigation within these middle layers, instead of manipulating attention weights, we introduce a Hidden State Rotation strategy. To explicitly correct the deviation, we rotate the hidden state $h_k$ toward the plane spanned by $C_T$ and $C_V$. Crucially, this rotation is norm-preserving, ensuring that the erroneous semantic direction is steered back to a reliable context without compromising the token's inherent representation strength. Furthermore, for the Transformer layers subsequent to the middle layers, we introduce a smoothing mechanism to maintain stability. For each layer in this later stage, we update the representation by performing a weighted interpolation between the current hidden state and the output from the preceding layer using a controllable smoothing factor. This ensures that the corrected semantic trajectory smoothly propagates to the final predictive layers.

Extensive experiments on widely adopted benchmarks (e.g., CHAIR, POPE, and MME) demonstrate the superiority of our proposed method across various LVLM architectures. In summary, our main contributions are as follows:
\begin{itemize}
    \item We propose a novel perspective for mitigating LVLM hallucinations by shifting the intervention focus from attention weights to hidden state vectors, addressing the semantic deviation directly in the hidden space.
    \item We introduce a robust hallucination detection mechanism based on hidden state dynamics. By observing that the semantic deviation indicator $E$ is predominantly high in the middle layers, we utilize a similarity-based threshold to precisely detect hallucinations in these susceptible intermediate stages.
    \item We design a layer-specific mitigation strategy. For deviating hidden states in the middle layers, we perform a norm-preserving rotation towards the plane formed by textual and visual context vectors; for subsequent layers, we employ a smoothing mechanism to maintain representational stability.
\end{itemize}

\section{Related Work}
\subsection{Large Vision-Language Models}
The rapid evolution of Large Language Models has catalyzed the development of Large Vision-Language Models, extending their profound reasoning capabilities into the multimodal domain. Models such as LLaVA \cite{liu2023visual} bridge the gap between modalities by aligning visual encoders with pre-trained language models through visual instruction tuning. Subsequent advancements, such as Qwen-VL \cite{bai2023qwen}, have further enhanced these capabilities by introducing robust fine-grained perception and supporting multi-image interactions at varying resolutions. Despite their remarkable success in diverse applications ranging from visual question answering to complex reasoning, these models predominantly suffer from a critical vulnerability: they frequently generate plausible yet ungrounded text, a phenomenon widely recognized as hallucination.

\subsection{Object Hallucination in LVLMs}
Object hallucination in LVLMs refers to the generation of objects, attributes, or relationships that are completely inconsistent with the provided visual input. Previous studies indicate that this flaw primarily stems from an over-reliance on statistical language priors, inadequate visual-linguistic alignment, and the inherent uncertainty during autoregressive decoding. To systematically quantify this issue, several authoritative benchmarks have been established. The CHAIR benchmark \cite{rohrbach2018object} evaluates hallucinations in open-ended image captioning by analyzing the proportion of generated ungrounded objects at both the instance and sentence levels. More recently, the POPE benchmark \cite{li2023evaluating} was introduced to assess the phenomenon in a more controlled, discriminative manner by framing the evaluation as a binary classification task regarding object existence. These metrics have become the standard testbed for evaluating the reliability of LVLMs.

\subsection{Training-Free Inference Intervention}
To mitigate hallucinations, researchers have explored both training-based and inference-time strategies. Given the prohibitive computational cost of retraining or RLHF alignment for massive LVLMs, training-free inference interventions have emerged as a highly efficient and scalable alternative. Recent methods primarily fall into three paradigms. Visual Contrastive Decoding \cite{leng2024mitigating} mitigates the over-reliance on language priors by contrasting the output distributions between original and distorted visual inputs, though it fundamentally doubles the inference latency. Attention manipulation approaches, such as OPERA \cite{huang2024opera}, introduce an over-trust penalty during beam search to trigger rollbacks when the model exhibits abnormal attention patterns. Feature injection methods, like MemVR \cite{zou2024look}, attempt to combat visual amnesia by performing memory-space visual retracing, which overwrites intermediate hidden states with key-value representations.

While these methods have demonstrated considerable success, they primarily rely on passive logits contrast, indirect attention manipulation, or rigid state substitution. In contrast, our proposed ROT framework pioneers a fundamentally different perspective. By discovering that hallucinated tokens suffer from a severe directional deviation in the hidden semantic space, ROT is the first to introduce an active geometric intervention. Rather than merely penalizing weights or replacing tokens, ROT directly rectifies the generation trajectory via a norm-preserving rotation. This targeted directional correction effectively re-anchors the hidden state to the factual multimodal subspace while preserving the original representational intensity, offering a highly efficient and structurally elegant solution to LVLM hallucinations.

\section{Motivation and Empirical Analysis}
\subsection{Preliminaries of Hidden States in LVLMs}
Large Vision-Language Models typically adopt a Transformer-based architecture to process interleaved multimodal inputs. Formally, given an input sequence consisting of both textual and visual tokens, the hidden representations are iteratively updated through multiple Transformer layers. In this work, we focus on the internal representation dynamics during the auto-regressive decoding phase. 

At a specific layer $k$, let $X^{k-1}$ denote the sequence of hidden states output from the preceding layer $k-1$. During the generation of the current token, the self-attention mechanism computes the attention scores between the current query and all preceding key representations. To analyze the information routed from different modalities, we extract the raw attention weights assigned to all past tokens and partition them into two distinct sets based on token modality: the textual attention weights and the visual attention weights. To accurately capture the aggregated semantic influence from each modality, we independently normalize these partitioned attention weights, yielding $\tilde{A}_{t,j}$ for the $j$-th text token and $\tilde{A}_{v,i}$ for the $i$-th vision token.

Using these normalized attention weights, we isolate the multimodal context by computing the weighted sum of the corresponding hidden states from $X^{k-1}$. Specifically, we construct a textual context vector $C_T$ and a visual context vector $C_V$, which are formulated as follows:
\begin{equation}
    \resizebox{0.88\linewidth}{!}{%
        $\displaystyle C_T = \sum_{j \in T} \tilde{A}_{t,j} X_j^{k-1}, \quad C_V = \sum_{i \in V} \tilde{A}_{v,i} X_i^{k-1}$%
    }
\end{equation}
where $T$ and $V$ represent the sets of indices for the preceding textual and visual tokens, respectively. The resulting vectors, $C_T$ and $C_V$, represent the isolated semantic subspaces grounded strictly in the textual and visual contexts.

Furthermore, we define $h_k$ as the hidden state vector of the currently generated token at layer $k$, extracted after the self-attention mechanism and the subsequent residual addition, but prior to the layer normalization operation. Because $h_k$ aggregates all contextual information and directly dictates the subsequent predictive trajectory through the Feed-Forward Network, investigating its geometric relationship with the modality-specific context vectors ($C_T$ and $C_V$) provides a fundamental lens into the semantic dynamics of the model, particularly when hallucinations occur.

\subsection{The Contextual Deviation of Hallucinated Tokens}
To systematically investigate the representation dynamics of hallucination, we conduct an extensive empirical study. We utilize the widely adopted LLaVA-1.5-7B model and decode 3,000 evaluation instances from the standard CHAIR benchmark. Following the official CHAIR evaluation protocol, we meticulously categorize the generated tokens into two distinct groups: normal tokens and hallucinated tokens. For each token at a given layer $k$, we compute its cosine similarity with the textual context vector ($S_T = \cos(h_k, C_T)$) and the visual context vector ($S_V = \cos(h_k, C_V)$). We then average these similarity scores across the two token groups for comparative visualization.

Under the prevailing language prior hypothesis, one might naturally expect hallucinated tokens to exhibit an over-reliance on linguistic context, which would mathematically manifest as an anomalously high similarity $S_T$ between the hidden state $h_k$ and the textual context $C_T$. However, our analysis reveals a striking, counter-intuitive phenomenon. 

\begin{figure}[htbp]
    \centering
    \begin{subfigure}{0.49\columnwidth}
        \includegraphics[width=\linewidth]{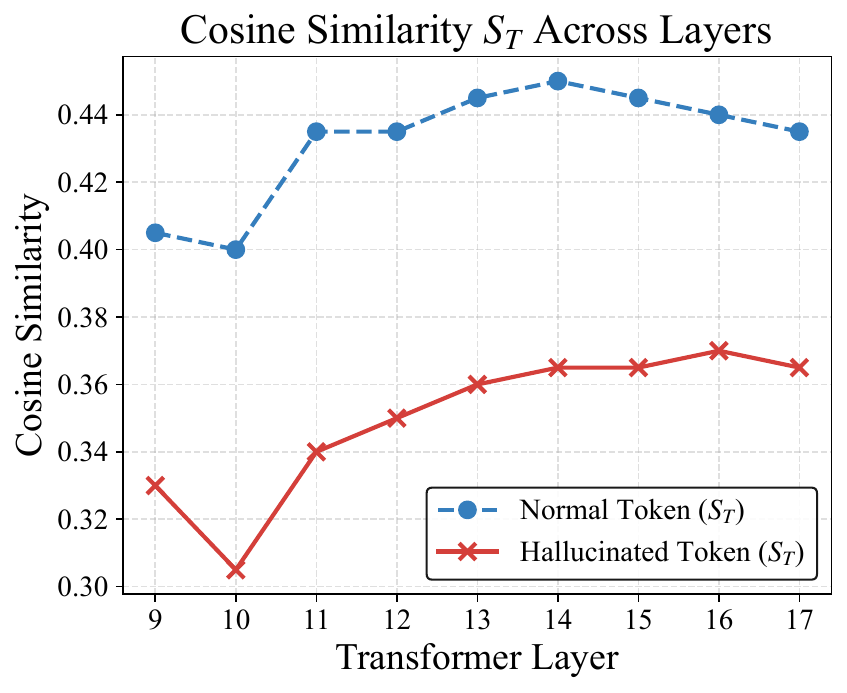}
    \end{subfigure}
    \hfill
    \begin{subfigure}{0.49\columnwidth}
        \includegraphics[width=\linewidth]{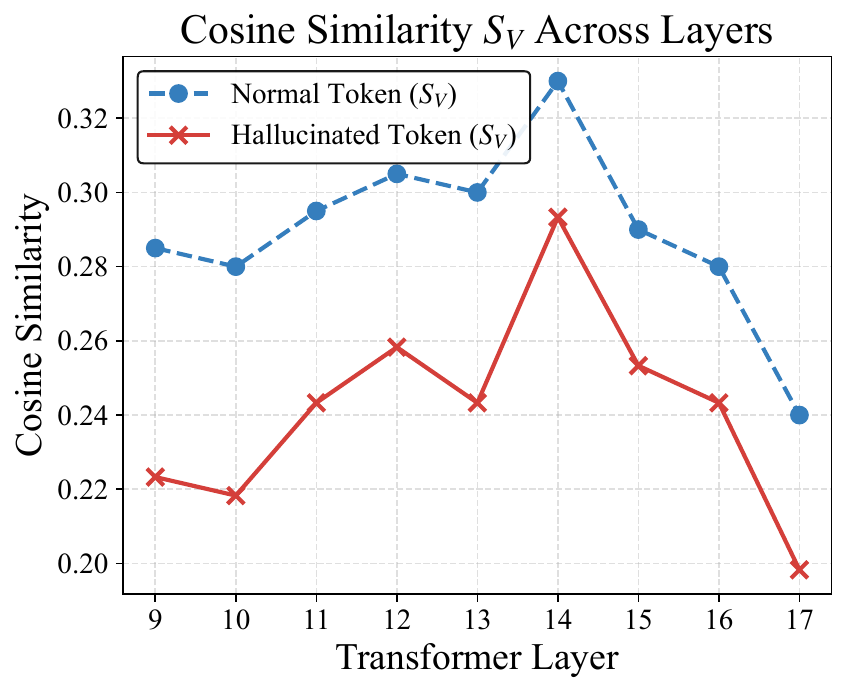}
    \end{subfigure}
    \caption{Average cosine similarities of the hidden state $h_k$ with the textual context ($S_T$) and visual context ($S_V$). Hallucinated tokens consistently exhibit significantly lower similarities to both contexts compared to normal tokens.}
    \label{fig:contextual_deviation}
\end{figure}

As illustrated in Figure \ref{fig:contextual_deviation}, the curves demonstrate a persistent and significant margin between normal and hallucinated tokens. Crucially, the representations of hallucinated tokens exhibit substantially lower cosine similarities with both the textual context $C_T$ and the visual context $C_V$ compared to their factual counterparts.

This observation profoundly refutes the simplistic language prior assumption. The simultaneous and sharp decline in both $S_T$ and $S_V$ indicates that the hidden state $h_k$ is semantically derailed. Rather than over-fitting to one modality, the hallucinated representation completely diverges from the grounded multimodal subspace spanned by $C_T$ and $C_V$, straying into an erroneous direction in the high-dimensional vector space.

This geometric divergence serves as the fundamental motivation for our method. It implies that passively scaling attention weights is insufficient; instead, an active geometric intervention is required to explicitly steer the diverging hidden states back to the grounded contextual subspace.

\section{Methodology}
\subsection{Dynamic Hallucination Detection}
Building upon the empirical observations in Section 3.2, we propose a dynamic detection mechanism to identify hallucinated tokens during the inference phase. Our goal is to quantify the semantic deviation of the hidden state $h_k$ relative to the grounded contextual subspace. To this end, we define a contextual deviation indicator, denoted as $E$, which aggregates the cosine distances between the current hidden state and the modality-specific context vectors:
\begin{equation}
    E = (1 - S_T) + (1 - S_V) = 2 - S_T - S_V
\end{equation}
where $S_T$ and $S_V$ represent the cosine similarities of $h_k$ with the textual context $C_T$ and visual context $C_V$, respectively. A higher value of $E$ indicates weaker joint alignment with the two modality-specific context anchors. We therefore use $E$ as a contextual-deviation risk indicator for intervention triggering, rather than as a calibrated binary classifier that necessarily identifies hallucination.

\begin{figure}[htbp]
    \centering
    \includegraphics[width=1.0\columnwidth]{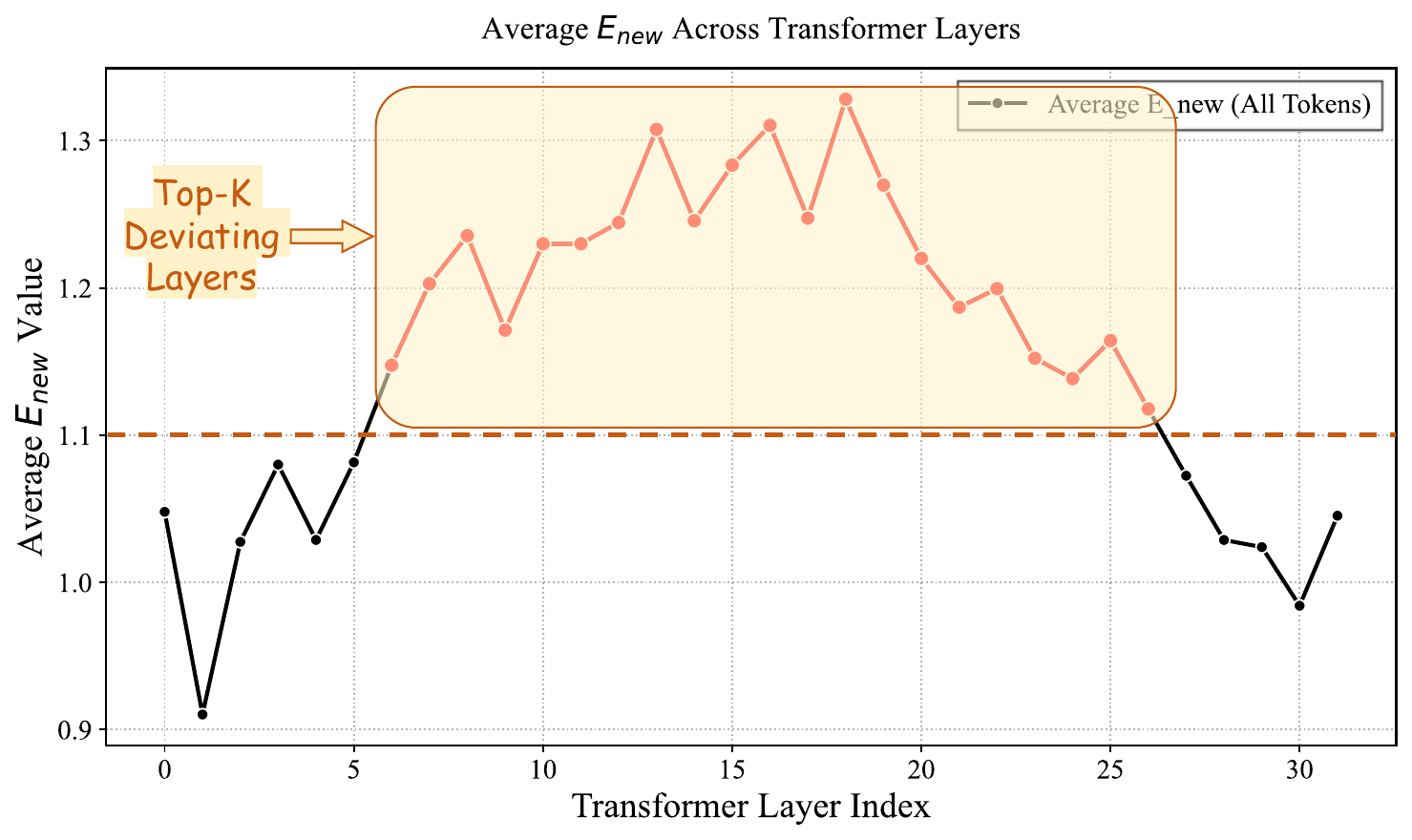}
    \caption{Distribution of the contextual deviation indicator $E$ across different transformer layers for LLaVA-1.5-7B. The deviation is significantly more pronounced in the intermediate layers.}
    \label{fig:layerwise_deviation}
\end{figure}

As illustrated in Figure \ref{fig:layerwise_deviation}, we quantify the average $E$ values across all layers using the LLaVA-1.5-7B model on the CHAIR benchmark. We observe a clear hierarchical aggregation effect: the semantic deviation is not uniformly distributed but is instead concentrated within specific depths of the transformer stack. Specifically, the indicator $E$ remains relatively low in the early and final layers but exhibits a significant peak in the intermediate stages. This suggests that the middle layers serve as a critical transition zone where multimodal features are fused, and it is within this zone that the hidden states are most susceptible to derailment.

\subsubsection{Top-K Deviating Layers Selection}
Based on the observed layer-wise distribution of $E$, we implement a selective intervention strategy rather than applying corrections to the entire network. While this process involves identifying a subset of layers, our approach is fundamentally driven by the magnitude of semantic divergence rather than a fixed layer count. Empirically, for models like LLaVA-1.5-7B, we identify the critical middle layers where the average deviation $E$ exceeds a specific sensitivity threshold. In our experimental setup, this criterion corresponds to layers 6 through 26, which we designate as the active execution zone for our proposed intervention.

Within these selected intermediate layers, we further refine the intervention to be token-adaptive by introducing a trigger threshold $\tau$. For each generated token, the Norm-Preserving Rotation is activated if and only if the current deviation indicator satisfies $E \geq \tau$. This dual-level selection mechanism ensures that the intervention is both spatially targeted at susceptible layers and dynamically triggered for specific tokens, effectively rectifying the generation trajectory only when a significant semantic derailment is detected. Such a design maintains the model's original representational integrity for factual tokens while precisely capturing and correcting those undergoing contextual deviation.

\subsection{Layer-Specific Hidden State Mitigation}
\subsubsection{Norm-Preserving Rotation (Top-K Deviating Layers)}
For the intermediate layers identified in the selection phase, we execute a geometric intervention termed Norm-Preserving Rotation to explicitly rectify the deviating hidden states. Our objective is to steer the derailed representation $h_k$ back toward the grounded contextual subspace while preserving its inherent representational intensity. 

First, we define a grounded semantic plane $\mathcal{P}$ formed by the textual and visual context vectors:
\begin{equation}
\mathcal{P} = \text{span}\{C_T, C_V\}
\end{equation}
This plane characterizes a factual subspace within the high-dimensional hidden space, providing a grounded reference for rectifying the deviating $h_k$ through rotation. Within this coplanar space, we establish a reference target direction, $C_{target}$, which represents the ideal grounded orientation for the current token. The orientation of $C_{target}$ is dynamically determined by balancing the textual and visual anchors:
\begin{equation}
    C_{target} = \text{Norm}((1-\alpha) \cdot C_T + \alpha \cdot C_V)
\end{equation}
where $\text{Norm}(\cdot)$ denotes the vector normalization operation, and $\alpha \in [0,1]$ is a hyperparameter that adjusts the relative proximity of the target direction toward the visual versus textual context.

Subsequently, we perform a rotation to move the hidden state $h_k$ toward $C_{target}$. To ensure that the intervention only modifies the semantic direction without disrupting the model's activation magnitude, we implement a norm-preserving rotation. The updated hidden state $h_{new}$ is formulated using the principle of spherical linear interpolation:
\begin{equation}
    h_{new} = \|h_k\| \cdot \text{Norm}\left( \text{Slerp}(h_k, C_{target}; \beta) \right)
\end{equation}
where $\beta \in [0,1]$ controls the rotation intensity. A larger $\beta$ signifies a stronger correction that pulls $h_k$ closer to the grounded plane. By maintaining $\|h_{new}\| = \|h_k\|$, we guarantee that the calibration does not inadvertently weaken the feature's signal strength before it enters the LayerNorm and the subsequent FFN. This precise directional correction ensures that the generation trajectory is re-anchored to the factual multimodal context.

\subsubsection{Representational Smoothing (Subsequent Layers)}
While the rotation mechanism effectively rectifies peak semantic deviations in the middle layers, applying aggressive directional changes in the final layers of the Transformer stack may interfere with the stabilization of the probability distribution over the vocabulary. Therefore, for the layers subsequent to the active execution zone, we transition from rotation to a Representational Smoothing strategy.

\begin{table*}[h]
    \centering
    \caption{Performance comparison of ROT against various state-of-the-art training-free baselines on LLaVA-1.5-7B across CHAIR, POPE, and MME benchmarks.}
    \label{tab:main_results_llava}
    \resizebox{\textwidth}{!}{
    \begin{tabular}{l cccccccccc}
        \toprule
        & \multicolumn{4}{c}{\textbf{CHAIR}} & \multicolumn{2}{c}{\textbf{POPE}} & \multicolumn{4}{c}{\textbf{MME}} \\
        \cmidrule(lr){2-5} \cmidrule(lr){6-7} \cmidrule(lr){8-11}
        \multirow{-2}{*}{\textbf{Method}} & $C_S\downarrow$ & $C_I\downarrow$ & Recall$\uparrow$ & length$\uparrow$ & Acc$\uparrow$ & F1$\uparrow$ & Exist.$\uparrow$ & Pos.$\uparrow$ & Color$\uparrow$ & Total$\uparrow$ \\
        \midrule
        Greedy & 54.4 & 15.7 & 75.2 & 82.7 & 85.5 & 85.9 & 175.67 & 114.00 & 151.00 & 440.67 \\
        VCD \cite{leng2024mitigating} & 51.1 & 14.7 & 75.2 & 82.3 & 85.0 & 85.3 & 184.67 & 128.67 & 153.00 & 466.34 \\
        OPERA \cite{huang2024opera} & 47.0 & 14.6 & 74.5 & 77.1 & 85.2 & 84.2 & 180.67 & 111.67 & 123.33 & 415.67 \\
        RITUAL \cite{woo2024ritual} & 45.2 & 13.2 & 74.3 & 80.3 & 84.3 & 85.2 & 187.50 & 125.00 & 164.17 & 476.67 \\
        Vissink \cite{kang2025see} & 52.4 & 14.5 & 75.1 & \underline{83.4} & 86.5 & 86.0 & 190.00 & 138.33 & 155.00 & 483.33 \\
        SID \cite{huo2024self} & 44.2 & 12.8 & 73.0 & 80.4 & 85.8 & 85.6 & 183.90 & 127.80 & 155.90 & 467.60 \\
        TAME \cite{tang2025intervening} & 41.3 & 12.4 & 74.4 & 79.9 & 85.7 & 85.4 & \underline{193.00} & \underline{139.00} & 164.67 & 496.67 \\
        LVLMs-Saliency \cite{zhang2026hallucination} & \underline{35.6} & \underline{8.2} & \underline{75.4} & 79.5 & \textbf{87.5} & \textbf{87.0} & 190.00 & 133.33 & \textbf{175.00} & \underline{498.33} \\
        \rowcolor{ourgreen}
        \textbf{ROT (Ours)} & \textbf{33.0} & \textbf{6.7} & \textbf{75.6} & \textbf{85.1} & \underline{86.6} & \underline{86.3} & \textbf{195.00} & \textbf{143.33} & \underline{168.33} & \textbf{506.66} \\
        \bottomrule
    \end{tabular}
    }
\end{table*}

\begin{table}[h]
    \centering
    \caption{Generalization performance of ROT across various LVLM architectures and parameter scales on CHAIR and the POPE benchmark (including its adversarial, popular, and random subsets).}
    \label{tab:generalization_results}
    \footnotesize
    \resizebox{\columnwidth}{!}{
    \begin{tabular}{l cccccccc}
        \toprule
        & \multicolumn{2}{c}{\textbf{CHAIR}} & \multicolumn{6}{c}{\textbf{POPE}} \\
        \cmidrule(lr){2-3} \cmidrule(lr){4-9}
        
        & & & \multicolumn{2}{c}{\textbf{Adversarial}} & \multicolumn{2}{c}{\textbf{Popular}} & \multicolumn{2}{c}{\textbf{Random}} \\
        \cmidrule(lr){4-5} \cmidrule(lr){6-7} \cmidrule(lr){8-9}
        
        \multirow{-3}{*}{\textbf{Method}} & $C_S\downarrow$ & $C_I\downarrow$ & Acc$\uparrow$ & F1$\uparrow$ & Acc$\uparrow$ & F1$\uparrow$ & Acc$\uparrow$ & F1$\uparrow$ \\
        \midrule
        
        \rowcolor{groupgray} 
        \multicolumn{9}{c}{\textbf{Base Model: LLaVA-1.5-7B}} \\
        Greedy & 54.4 & 15.7 & 80.4 & 81.7 & 86.2 & 86.4 & 89.9 & 89.6 \\
        VCD & 51.1 & 14.7 & 81.2 & 82.1 & 85.7 & 85.9 & 88.0 & 88.1 \\
        MemVR & 50.4 & 14.3 & 79.8 & 81.8 & 86.3 & 85.9 & 89.4 & 89.6 \\
        \rowcolor{ourgreen} \textbf{ROT} & \textbf{33.0} & \textbf{6.7} & \textbf{82.3} & \textbf{82.6} & \textbf{87.2} & \textbf{86.6} & \textbf{90.2} & \textbf{89.8} \\
        \midrule
        
        \rowcolor{groupgray} 
        \multicolumn{9}{c}{\textbf{Base Model: LLaVA-1.5-13B}} \\
        Greedy & 41.4 & 12.7 & 83.8 & 84.4 & 86.1 & 86.8 & 87.5 & 88.4 \\
        VCD & 44.7 & 13.1 & 81.9 & 81.8 & 85.7 & 85.1 & 87.4 & 86.6 \\
        MemVR & 42.2 & 11.5 & 81.2 & 82.4 & 87.1 & 85.3 & 88.2 & 88.5 \\
        \rowcolor{ourgreen} \textbf{ROT} & \textbf{30.8} & \textbf{4.3} & \textbf{85.0} & \textbf{84.5} & \textbf{87.7} & \textbf{86.9} & \textbf{89.5} & \textbf{88.6} \\
        \midrule
        
        \rowcolor{groupgray} 
        \multicolumn{9}{c}{\textbf{Base Model: Qwen2-VL-7B}} \\
        Greedy & 24.2 & 8.0 & 85.9 & 84.8 & 87.6 & 86.4 & 89.3 & 88.6 \\
        \rowcolor{ourgreen} \textbf{ROT} & \textbf{21.4} & \textbf{4.9} & \textbf{87.3} & \textbf{86.2} & \textbf{89.2} & \textbf{88.5} & \textbf{90.5} & \textbf{89.9} \\
        \midrule
        
        \rowcolor{groupgray} 
        \multicolumn{9}{c}{\textbf{Base Model: Qwen2.5-VL-32B}} \\
        Greedy & 43.6 & 9.5 & 85.8 & 84.6 & 86.9 & 85.6 & 88.5 & 87.1 \\
        \rowcolor{ourgreen} \textbf{ROT} & \textbf{30.9} & \textbf{8.7} & \textbf{86.5} & \textbf{85.5} & \textbf{87.4} & \textbf{86.4} & \textbf{89.3} & \textbf{88.1} \\
        \bottomrule
    \end{tabular}
    }
\end{table}

In this stage, rather than forcing a geometric redirection, we aim to enhance the coherence of the corrected trajectory by incorporating historical semantic information. For each layer $k$ in the later stages, the hidden state is updated via a weighted interpolation with the representation from the preceding layer:
\begin{equation}
    h_k = \gamma h_{k-1} + (1-\gamma) h_k
\end{equation}
where $\gamma \in [0,1]$ is the smoothing factor. This operation facilitates a more consistent representation update, allowing the calibrated semantic signals from the intermediate layers to propagate reliably to the final prediction head. This dual-stage intervention, which combines rotation for correction and smoothing for stability, ensures a robust and grounded generation process throughout the entire network depth.

\section{Experiments}
\subsection{Experimental Setups}
\paragraph{Models.} To comprehensively evaluate the effectiveness and generalizability of our proposed ROT framework, we conduct experiments across multiple mainstream Large Vision-Language Models with varying architectures and parameter scales. Specifically, we select the widely adopted LLaVA-1.5 series, including LLaVA-1.5-7B and LLaVA-1.5-13B \cite{liu2024improved}. Furthermore, to validate our method on state-of-the-art models, we incorporate the Qwen-VL family, utilizing both Qwen2-VL-7B \cite{wang2024qwen2} and the highly capable Qwen2.5-VL-32B \cite{qwen2.5vl2025}.

\paragraph{Baselines.} We compare ROT against standard Greedy Decoding and a comprehensive suite of strong inference baselines. These include contrastive-based methods such as VCD \cite{leng2024mitigating}, attention-penalty methods like OPERA \cite{huang2024opera}, and recent advanced mitigation strategies including RITUAL \cite{woo2024ritual}, Vissink \cite{kang2025see}, SID \cite{huo2024self}, and TAME \cite{tang2025intervening}. Additionally, we compare our method with LVLMs-Saliency \cite{zhang2026hallucination}, a recent gradient-aware sampling approach.

\paragraph{Benchmarks and Metrics.} To evaluate the effectiveness of our approach in mitigating object hallucinations, we conduct experiments on three widely adopted benchmarks: CHAIR \cite{rohrbach2018object}, POPE \cite{li2023evaluating}, and MME \cite{fu2026mme}. Following standard evaluation protocols, we primarily evaluate the hallucination rate at the sentence level $C_S$ and instance level $C_I$ for the CHAIR benchmark, where lower scores indicate better hallucination mitigation.

\subsection{Main Results}
\paragraph{Mitigation of Object Hallucination.} We first evaluate the hallucination mitigation performance on the LLaVA-1.5-7B model, as detailed in Table \ref{tab:main_results_llava}. On the CHAIR benchmark, our ROT framework achieves remarkable quantitative improvements, establishing a new standard for training-free interventions. Compared to the standard greedy decoding, ROT drastically reduces the sentence-level hallucination metric $C_S$ from 54.4 to 33.0 and the instance-level metric $C_I$ from 15.7 to 6.7. Even when compared to the highly competitive, second-best method, LVLMs-Saliency, our geometric intervention proves superior, further depressing $C_S$ by an absolute margin of 2.6 points and $C_I$ by 1.5 points. Transitioning to the POPE benchmark, ROT continues to deliver robust quantitative gains. It elevates the Accuracy from the greedy baseline of 85.5 to 86.6 and boosts the F1-score from 85.9 to 86.3, decisively outperforming widely adopted baseline strategies such as OPERA, VCD, and TAME. On the comprehensive MME benchmark, ROT surpasses all evaluated methods and achieves a state-of-the-art total score of 506.66. This validates that correcting the semantic trajectory directly in the hidden space is a highly effective mechanism for anchoring generation to factual visual evidence across diverse evaluation protocols.

\paragraph{Generalization Across Architectures and Scales.} To demonstrate the universal applicability and scalability of ROT, we extend our evaluation across diverse model families and parameter sizes, encompassing LLaVA-1.5-7B, LLaVA-1.5-13B, Qwen2-VL-7B, and Qwen2.5-VL-32B, as presented in Table \ref{tab:generalization_results}. Across all tested architectures, ROT consistently and significantly suppresses hallucinated outputs on the CHAIR benchmark. Quantitatively, the intervention achieves an average relative reduction of 26.4\% in $C_S$ and an impressive 42.7\% in $C_I$ across the four models. Beyond CHAIR, we comprehensively assess factual grounding using the POPE benchmark. Across the four diverse models and all testing subsets, our method guarantees steady improvements, yielding consistent Accuracy lifts of up to 2.0 absolute points relative to their respective greedy decoding baselines. Exhibiting robust efficacy even on massive, highly optimized models like Qwen2.5-VL-32B underscores that the semantic deviation of hidden states is a fundamental issue across LVLMs, and our geometric intervention provides a scalable, architecture-agnostic solution.

\subsection{Ablation Study}
To validate the efficacy and necessity of the core components within our proposed framework, we conduct a comprehensive ablation study on the CHAIR benchmark. We isolate the two primary mechanisms of our approach: the Norm-Preserving Rotation in the intermediate layers (denoted as w/o Rotation) and the Representational Smoothing in the subsequent layers (denoted as w/o Smoothing). The evaluation is performed on both LLaVA-1.5-7B and Qwen2-VL-7B architectures, with the quantitative results presented in Table \ref{tab:ablation_results}.

Removing the rotation mechanism causes severe performance degradation. On LLaVA-1.5-7B, omitting rotation drastically increases $C_I$ from 6.7 to 12.4 and $C_S$ from 33.0 to 40.5. This confirms that merely applying passive smoothing is insufficient; actively redirecting the hidden state toward the grounded factual plane is essential to fundamentally correct semantic deviations.

Similarly, disabling the smoothing mechanism notably increases hallucination rates, degrading LLaVA-1.5-7B's $C_I$ from 6.7 to 8.3. This indicates that while rotation effectively anchors the semantic direction, the absence of subsequent smoothing allows the calibrated hidden states to destabilize in deeper layers, triggering secondary ungrounded fluctuations. 

Ultimately, the full ROT framework significantly outperforms both incomplete variants. The norm-preserving rotation provides crucial geometric correction at peak deviation, while smoothing reliably preserves this structural integrity to the output head, confirming their strong synergy in mitigating multimodal hallucinations.

\begin{table}[h]
    \centering
    \caption{Ablation study of ROT components on LLaVA-1.5-7B and Qwen2-VL-7B models across the CHAIR benchmark.}
    \label{tab:ablation_results}
    \footnotesize
    \begin{tabularx}{\columnwidth}{l YYYY}
        \toprule
        & \multicolumn{2}{c}{\textbf{LLaVA-1.5-7B}} & \multicolumn{2}{c}{\textbf{Qwen2-VL-7B}} \\
        \cmidrule(lr){2-3} \cmidrule(lr){4-5}
        \multirow{-2}{*}{\textbf{Method}} & $C_S\downarrow$ & $C_I\downarrow$ & $C_S\downarrow$ & $C_I\downarrow$ \\
        \midrule
        Greedy & 54.5 & 15.7 & 24.2 & 8.0 \\
        \textbf{ROT} & \textbf{33.0} & \textbf{6.7} & \textbf{21.4} & \textbf{4.9} \\
        w/o Rotation & 40.5 & 12.4 & 23.4 & 6.6 \\
        w/o Smoothing & 38.2 & 8.3 & 21.9 & 5.2 \\
        \bottomrule
    \end{tabularx}
\end{table}

\subsection{Case Study}
\begin{figure}[htbp]
    \centering
    \includegraphics[width=\columnwidth]{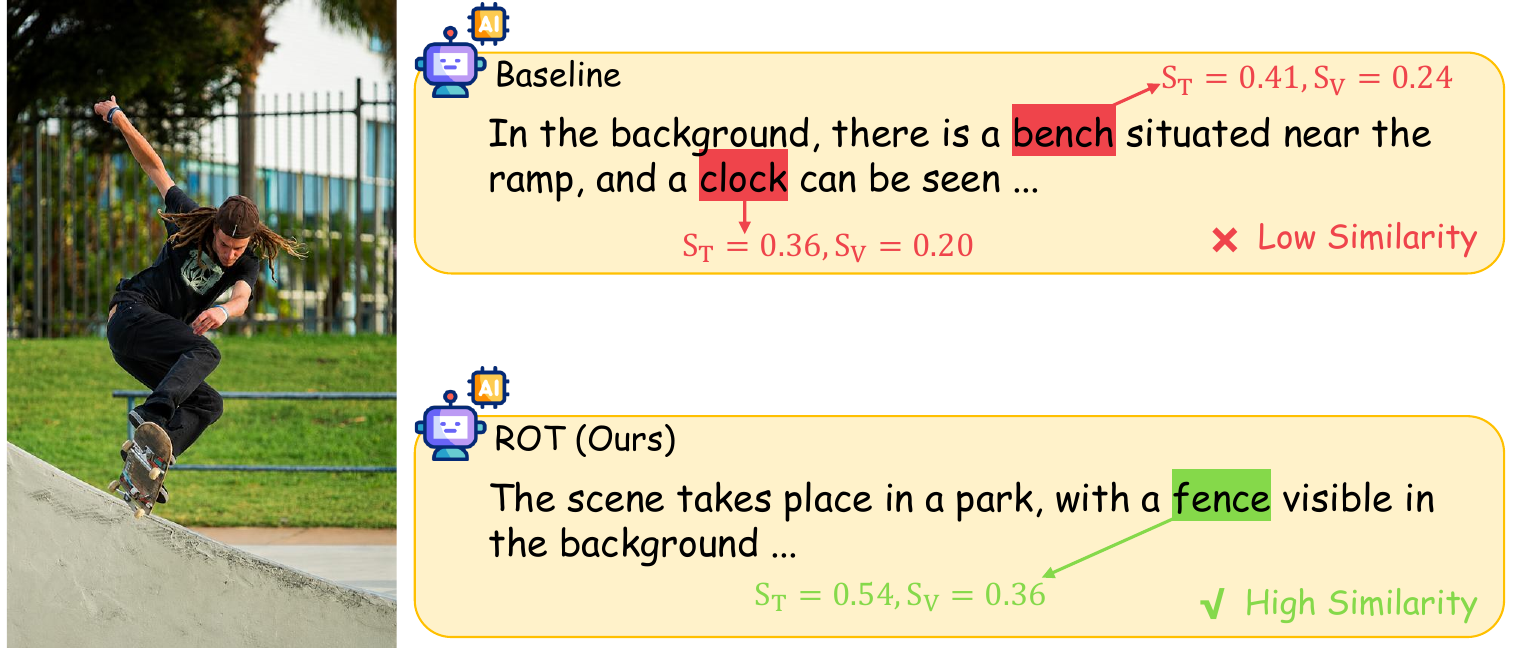}
    \caption{Qualitative comparison and hidden state similarities. The baseline model hallucinates a bench and a clock, which exhibit distinctly lower $S_T$ and $S_V$ compared to the factual token fence generated by ROT.}
    \label{fig:case_study}
\end{figure}

To intuitively demonstrate the mechanism and effectiveness of our proposed method, we present a qualitative case study from the CHAIR benchmark using the LLaVA-1.5-7B model, as illustrated in Figure \ref{fig:case_study}.

In this example, given an image of a person skateboarding, the baseline model generates a description containing ungrounded objects, specifically hallucinating a bench situated near the ramp and a clock on the wall. By analyzing the internal representations at the moment of generation, we find that these hallucinated tokens exhibit notably low contextual similarities. Specifically, the token bench presents an $S_T$ of 0.41 and $S_V$ of 0.24, while the token clock further drops to an $S_T$ of 0.36 and $S_V$ of 0.20. These values indicate that the hidden states of these tokens have deviated significantly from the grounded textual and visual context.

In contrast, our ROT framework dynamically detects this deviation and applies the norm-preserving rotation to steer the representation back toward the grounded semantic plane. As a result, the model successfully identifies the factual object in the scene, generating the token fence with significantly higher similarities ($S_T = 0.54$, $S_V = 0.36$). This case demonstrates that ROT effectively re-anchors the generation trajectory to the visual evidence, replacing ungrounded hallucinations with accurate, grounded descriptions.

\section{Conclusion}
In this paper, we introduce ROT, a novel training-free framework designed to mitigate object hallucinations in LVLMs. Our approach is motivated by the empirical discovery that hallucinated tokens exhibit a significant directional deviation in the hidden semantic space, rather than a simple over-reliance on language priors. To address this, we propose a layer-specific intervention strategy consisting of two key components: a norm-preserving rotation that explicitly corrects the semantic trajectory in the most deviating intermediate layers, and a representational smoothing mechanism that ensures stability in subsequent layers. Extensive experiments across multiple benchmarks, including CHAIR, POPE, and MME, demonstrate that ROT consistently reduces hallucination rates and outperforms various state-of-the-art baselines while maintaining the model's fundamental multimodal capabilities. Our findings underscore the importance of geometric calibration in the hidden space as a robust and efficient solution for grounded generation.

\section{Limitations}
Despite the effectiveness of the ROT framework, several limitations remain to be addressed in future work. First, the construction of the grounded semantic plane is derived solely from the local textual and visual contexts provided within the model's attention mechanism. While effective, this approach does not incorporate external structured knowledge or global scene graphs, which could potentially offer a more comprehensive reference for resolving complex hallucinations. Second, the identification of the critical deviating layers and the selection of the trigger threshold $\tau$ are currently based on empirical observations from representative model families. While ROT demonstrates strong generalization across different scales (7B to 32B), these configurations might require adaptive fine-tuning or automated search strategies when applied to ultra-large-scale models with significantly different depth-wise representational dynamics.

\bibliography{custom}

\nocite{*}

\newpage

\appendix

\section{Algorithm of the ROT Framework}
\label{sec:appendix_algorithm}

To facilitate a deeper understanding and ensure the reproducibility of our proposed method, we provide the detailed pseudocode for the ROT framework in Algorithm \ref{alg:rot_pseudocode}. 

The algorithm operates during the autoregressive decoding phase at each generation step. It takes the layer-wise hidden states and attention distributions as inputs, dynamically computes the modality-specific contexts, and applies the dual-stage intervention (Norm-Preserving Rotation and Representational Smoothing) based on the layer depth and the contextual deviation indicator $E$.

\begin{algorithm}[htbp]
\caption{Dynamic Hallucination Detection and Mitigation via ROT}
\label{alg:rot_pseudocode}
\textbf{Input:} Current token hidden states $X$; Normalized attention weights $\tilde{A}$; Textual token indices $T$; Visual token indices $V$. \\
\textbf{Parameters:} Rotation execution zone (layers $1$ to $L_{mid}$); Smoothing zone (layers $L_{mid}+1$ to $L$); Trigger threshold $\tau$; Balance factor $\alpha$; Rotation intensity $\beta$; Smoothing factor $\gamma$. \\
\textbf{Output:} Calibrated hidden states for vocabulary prediction.

\begin{algorithmic}[1]
\State Initialize $h_0$ as the input embedding of the current token.
\For{each transformer layer $k = 1$ to $L$}
    \State \textit{// Forward pass through Self-Attention}
    \State $\text{Attn\_Output}, \tilde{A} \leftarrow \text{Self-Attention}(h_{k-1}, X)$
    \State $h_k \leftarrow h_{k-1} + \text{Attn\_Output}$ \Comment{Extract post-residual hidden state}
    
    \State \textit{// Context Aggregation}
    \State $C_T \leftarrow \sum_{j \in T} \tilde{A}_{t,j} X_j^{k-1}$ \Comment{Textual context}
    \State $C_V \leftarrow \sum_{i \in V} \tilde{A}_{v,i} X_i^{k-1}$ \Comment{Visual context}
    
    \State \textit{// Layer-Specific Interventions}
    \If{$k \leq L_{mid}$} \Comment{Active execution zone for Rotation}
        \State $S_T \leftarrow \cos(h_k, C_T)$
        \State $S_V \leftarrow \cos(h_k, C_V)$
        \State $E \leftarrow 2 - S_T - S_V$ \Comment{Calculate contextual deviation}
        
        \If{$E \geq \tau$} \Comment{Trigger Norm-Preserving Rotation}
            \State $C_{target} \leftarrow \text{Norm}(\alpha \cdot C_V + (1-\alpha) \cdot C_T)$
            \State $h_{new} \leftarrow \|h_k\| \cdot \text{Norm}\left( \text{Slerp}(h_k, C_{target}; \beta) \right)$
            \State $h_k \leftarrow h_{new}$ \Comment{Rectify semantic trajectory}
        \EndIf
        
    \Else \Comment{Subsequent layers for Smoothing}
        \State $h_k \leftarrow \gamma \cdot h_{k-1}^{(prev)} + (1-\gamma) \cdot h_k$ \Comment{Representational smoothing}
    \EndIf
    
    \State $h_{k-1}^{(prev)} \leftarrow h_k$ \Comment{Cache current state for the next layer's smoothing}
    
    \State \textit{// Forward pass through FFN}
    \State $h_k \leftarrow \text{LayerNorm}(h_k)$
    \State $h_k \leftarrow h_k + \text{FFN}(h_k)$
\EndFor
\State \textbf{return} $h_L$ \Comment{Final hidden state for language modeling head}
\end{algorithmic}
\end{algorithm}

\section{Evaluation Metrics Definitions}
\label{sec:appendix_metrics}

To comprehensively evaluate the hallucination mitigation performance, we employ three widely used benchmarks. The detailed formulations of their respective metrics are defined as follows:

\paragraph{CHAIR Metrics.} The Caption Hash Algorithm for Image Reporting (CHAIR) benchmark quantifies object hallucination in open-ended image captioning at two different granularities: the instance level ($C_I$) and the sentence level ($C_S$). 
The instance-level metric $C_I$ calculates the proportion of hallucinated object instances among all objects mentioned by the model:
\begin{equation}
    C_I = \frac{\text{\# of hallucinated objects}}{\text{\# of all objects mentioned}}
\end{equation}
The sentence-level metric $C_S$ measures the proportion of sentences that contain at least one hallucinated object:
\begin{equation}
    C_S = \frac{\text{\# of sentences with hallucinated objects}}{\text{\# of all sentences}}
\end{equation}
For both metrics, a lower value indicates a reduced degree of object hallucination.

\paragraph{POPE Metrics.} The Polling-based Object Probing Evaluation (POPE) frames hallucination evaluation as a binary visual question answering task. We report the standard Accuracy (Acc) and F1-score:
\begin{itemize}
    \item \textbf{Accuracy (Acc):} The ratio of correctly answered probing questions (both existing and non-existing objects) to the total number of questions.
    \item \textbf{F1-score:} The harmonic mean of Precision and Recall. It provides a balanced measure of the model's ability to correctly identify existing objects without falsely confirming non-existent ones.
\end{itemize}

\paragraph{MME Metrics.} For the MME benchmark, we focus on the foundational perception subtasks (e.g., Existence, Position, Color). Following the official MME scoring protocol, the score for each subtask is derived from a combination of basic accuracy and pair-wise accuracy. Specifically, the evaluation utilizes pairs of positive and negative questions. The maximum possible score for each subtask is 200, where 100 points correspond to the basic accuracy across all individual questions, and the remaining 100 points are awarded based on strict pair-wise accuracy (i.e., both questions regarding a specific object must be answered correctly).

\section{Experimental Details}
\label{sec:appendix_experimental_details}

In this section, we elaborate on the concrete hardware environment and the specific hyperparameter configurations employed throughout our empirical evaluations to guarantee reproducibility. All experiments, encompassing both the standard greedy decoding baselines and our proposed ROT framework, are consistently executed on a unified computing node equipped with an NVIDIA A800 (80GB) GPU.

The dual-stage intervention mechanism of ROT is governed by three key hyperparameters: the context balance factor $\alpha$, the rotation intensity $\beta$, and the representational smoothing factor $\gamma$. Taking CHAIR as an example, for the LLaVA model family, the hyperparameters for the LLaVA-1.5-7B model are established at $\alpha = 0.44$, $\beta = 0.062$, and $\gamma = 0.51$. When scaling up to the larger LLaVA-1.5-13B architecture, the configurations are tuned to $\alpha = 0.45$, $\beta = 0.050$, and $\gamma = 0.55$. On the Qwen2-VL-7B model, we configure the intervention framework with $\alpha = 0.50$, $\beta = 0.048$, and $\gamma = 0.43$. For the Qwen2.5-VL-32B model, the parameters are fine-tuned to $\alpha = 0.46$, $\beta = 0.024$, and $\gamma = 0.41$.

\section{System Prompts}
\label{sec:appendix_prompts}

In our experiments, we design specific system prompts to guide the Large Vision-Language Models to generate appropriate responses tailored to the evaluation formats of different benchmarks. To ensure fair and rigorous comparisons, these instruction templates are kept strictly identical across the baseline methods and our proposed ROT framework.

For the open-ended image captioning task in the \textbf{CHAIR} benchmark, we utilize a standard descriptive prompt to encourage comprehensive textual generation:
\begin{quote}
    \texttt{"Please describe this image in detail."}
\end{quote}

For the probing-based tasks in the \textbf{POPE} and \textbf{MME} benchmarks, we instruct the model to first briefly analyze the image content and then output the final binary classification result. This step-by-step instruction effectively standardizes the output format for automated exact-match evaluation:
\begin{quote}
    \texttt{"Please describe the image content concisely, and then answer with yes or no at the end."}
\end{quote}

\section{Extensive Qualitative Results}
\label{sec:appendix_case_studies}

To further demonstrate the robust effectiveness of the ROT framework in mitigating object hallucinations, we present additional qualitative case studies in this section. These examples illustrate how our geometric intervention correctly anchors the generation trajectory across diverse and complex visual scenes.

\begin{figure}[H]
    \centering
    \begin{subfigure}{\columnwidth}
        \includegraphics[width=\linewidth]{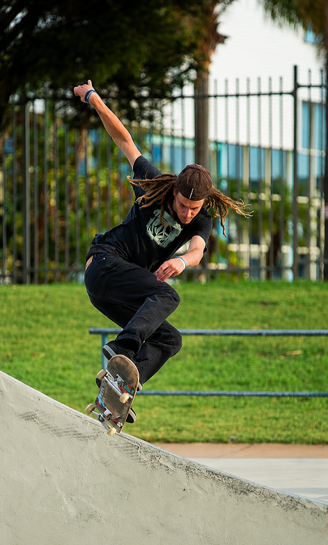}
    \end{subfigure}
    \begin{subfigure}{\columnwidth}
        \includegraphics[width=\linewidth]{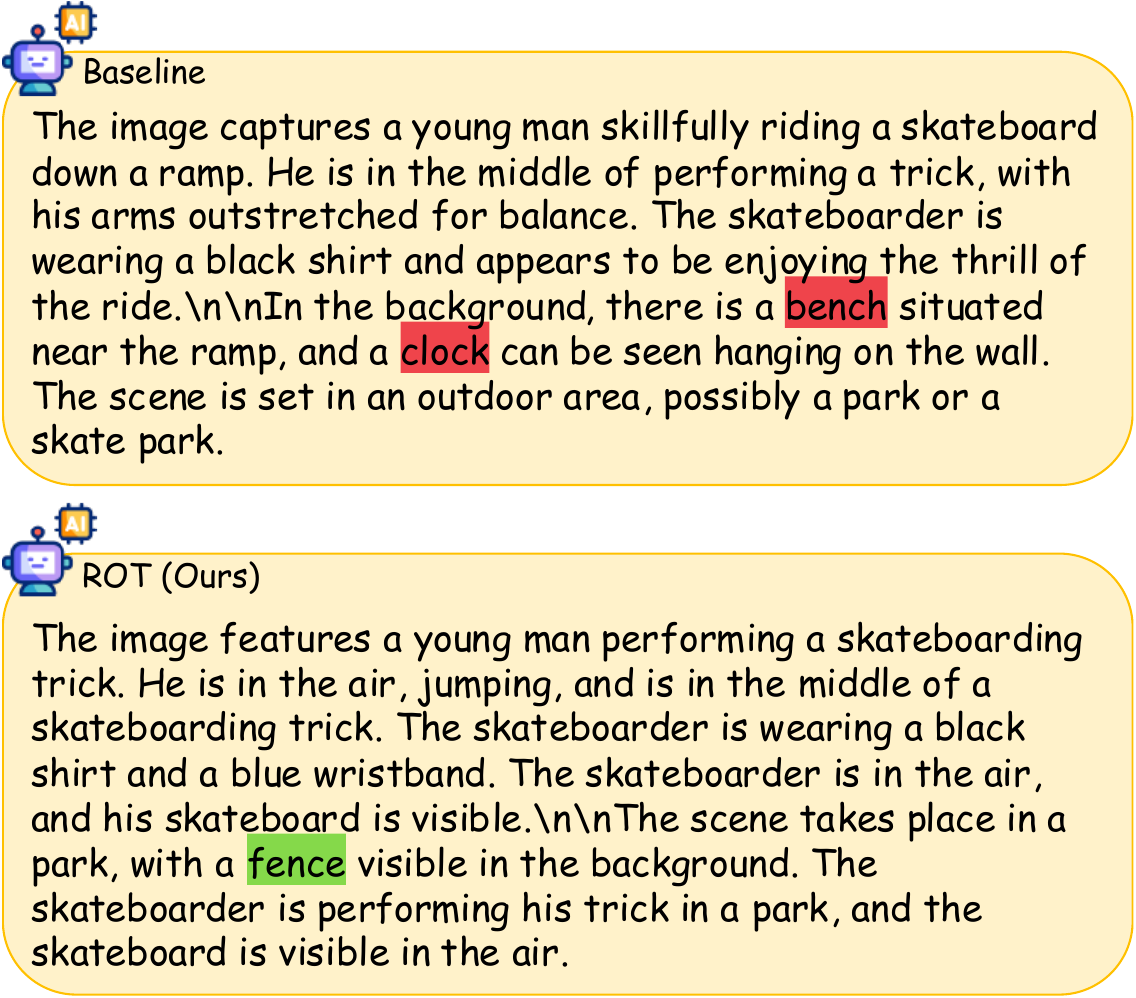}
    \end{subfigure}
    \caption{Qualitative Example 1. ROT successfully removes the hallucinated ``bench'' and ``clock'', and accurately describes the ``fence''.}
    \label{fig:appendix_case1}
\end{figure}

As illustrated in Figure \ref{fig:appendix_case1}, the standard greedy decoding hallucinates non-existent objects, specifically fabricating a ``bench'' and a ``clock''. Our proposed ROT method effectively corrects this contextual deviation, completely eliminating these hallucinations while successfully identifying and generating the correct factual token ``fence''.

\begin{figure}[H]
    \centering
    \begin{subfigure}{0.9\columnwidth}
        \includegraphics[width=\linewidth]{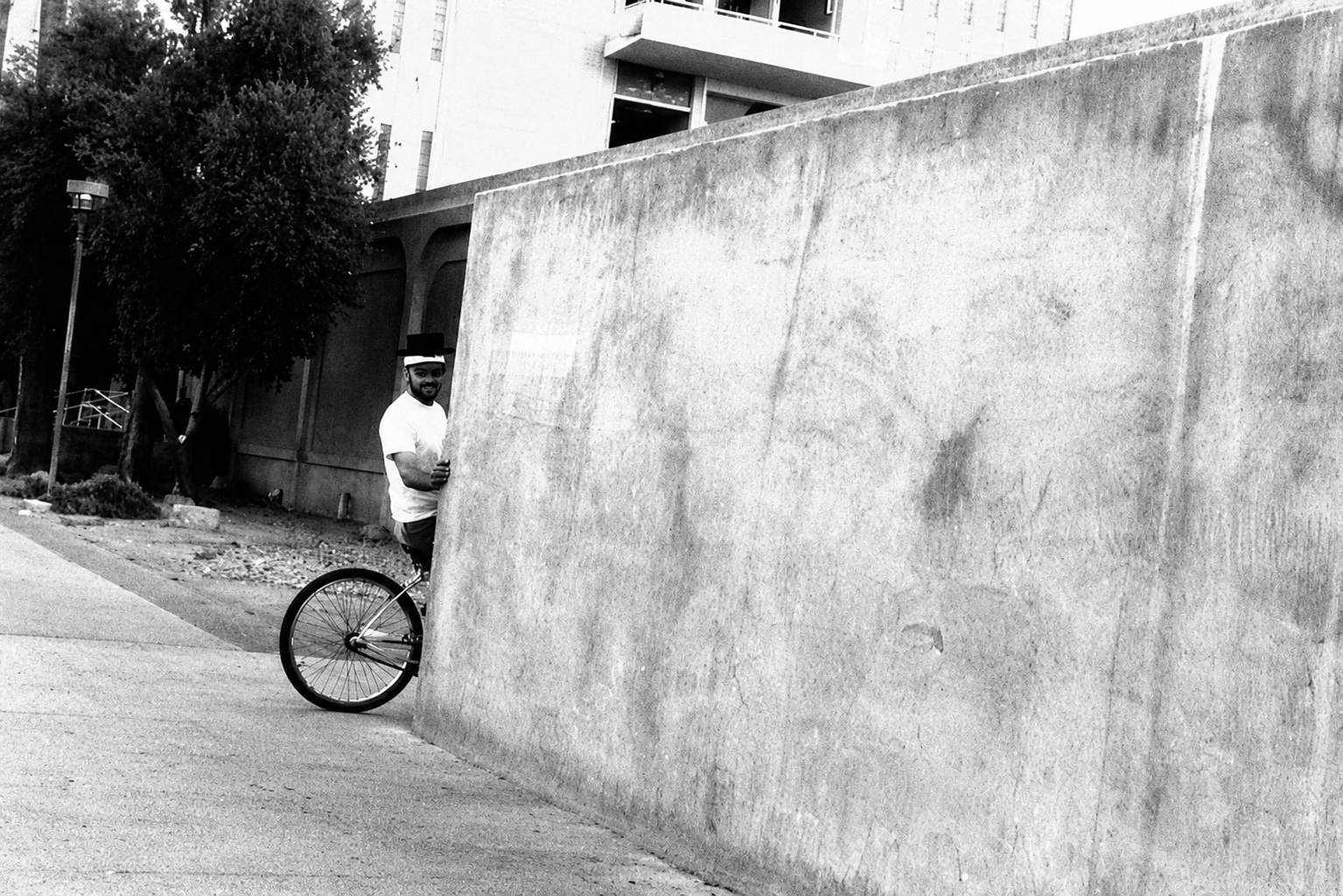}
    \end{subfigure}
    \begin{subfigure}{0.95\columnwidth}
        \includegraphics[width=\linewidth]{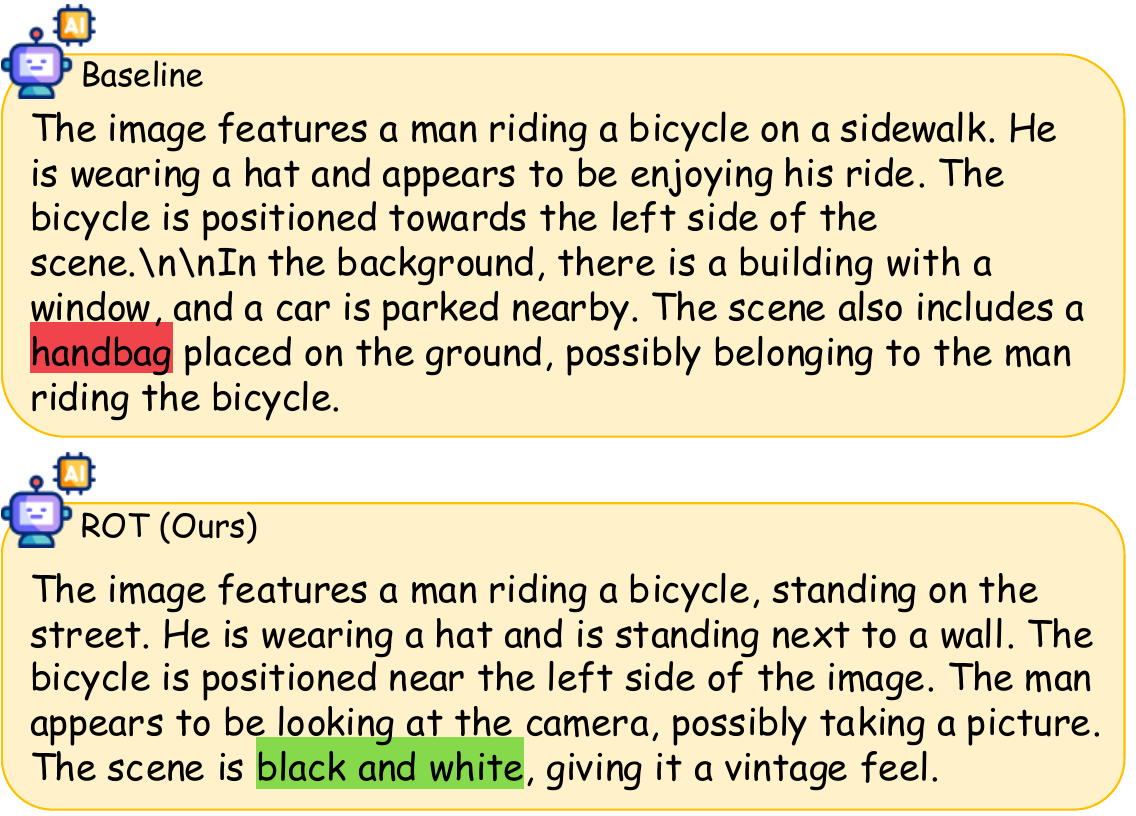}
    \end{subfigure}
    \caption{Qualitative Example 2. ROT eliminates the hallucinated ``handbag'' and correctly identifies the ``black and white'' attribute of the image.}
    \label{fig:appendix_case2}
\end{figure}

In Figure \ref{fig:appendix_case2}, the baseline model erroneously generates a ``handbag'' that does not exist in the visual input. The ROT framework not only removes this hallucinated object but also demonstrates a strong capability to capture global visual attributes, accurately describing the overall scene as ``black and white'' to match the monochrome nature of the image.

\begin{figure}[H]
    \centering
    \begin{subfigure}{0.9\columnwidth}
        \includegraphics[width=\linewidth]{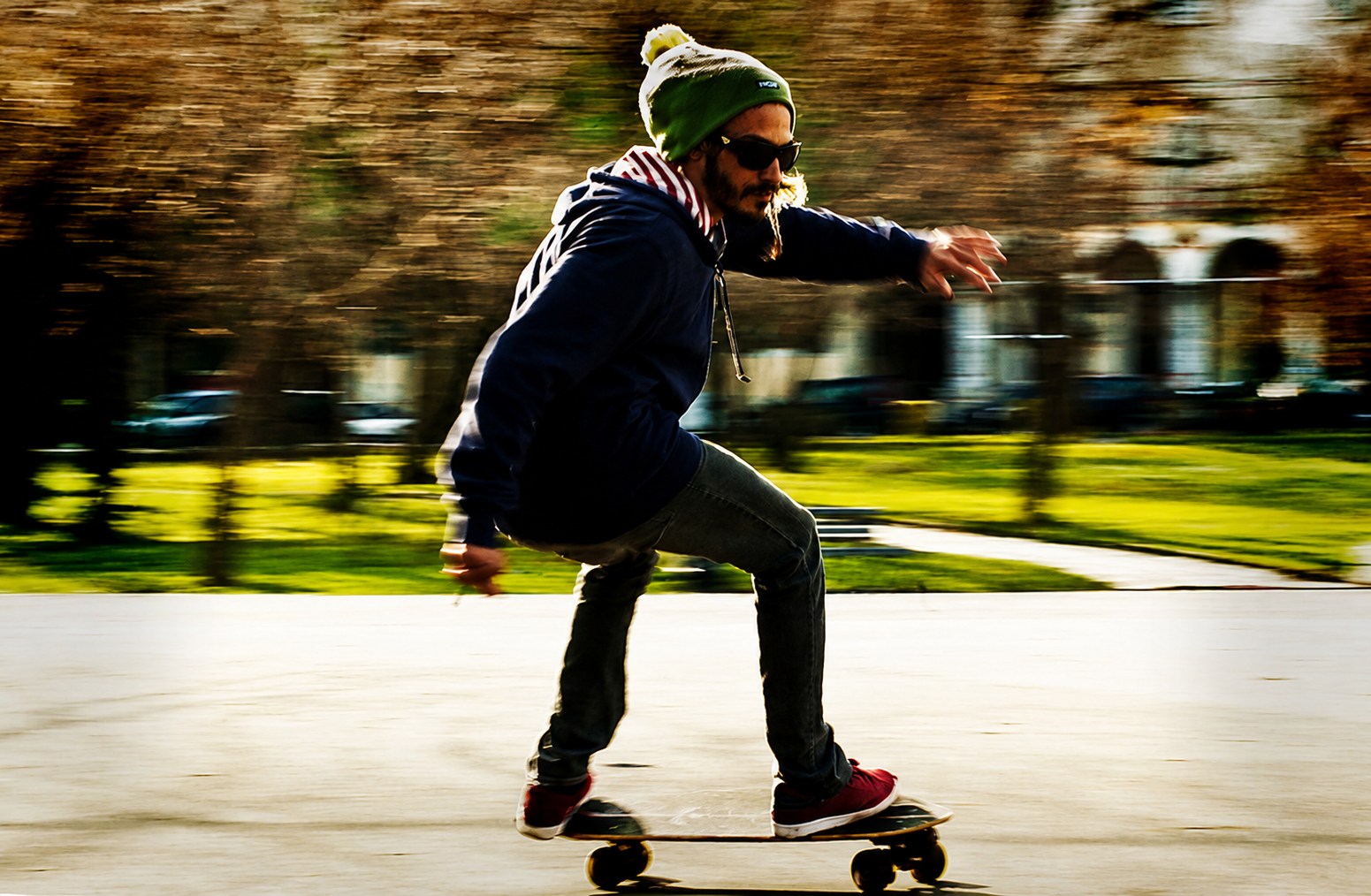}
    \end{subfigure}
    \begin{subfigure}{0.95\columnwidth}
        \includegraphics[width=\linewidth]{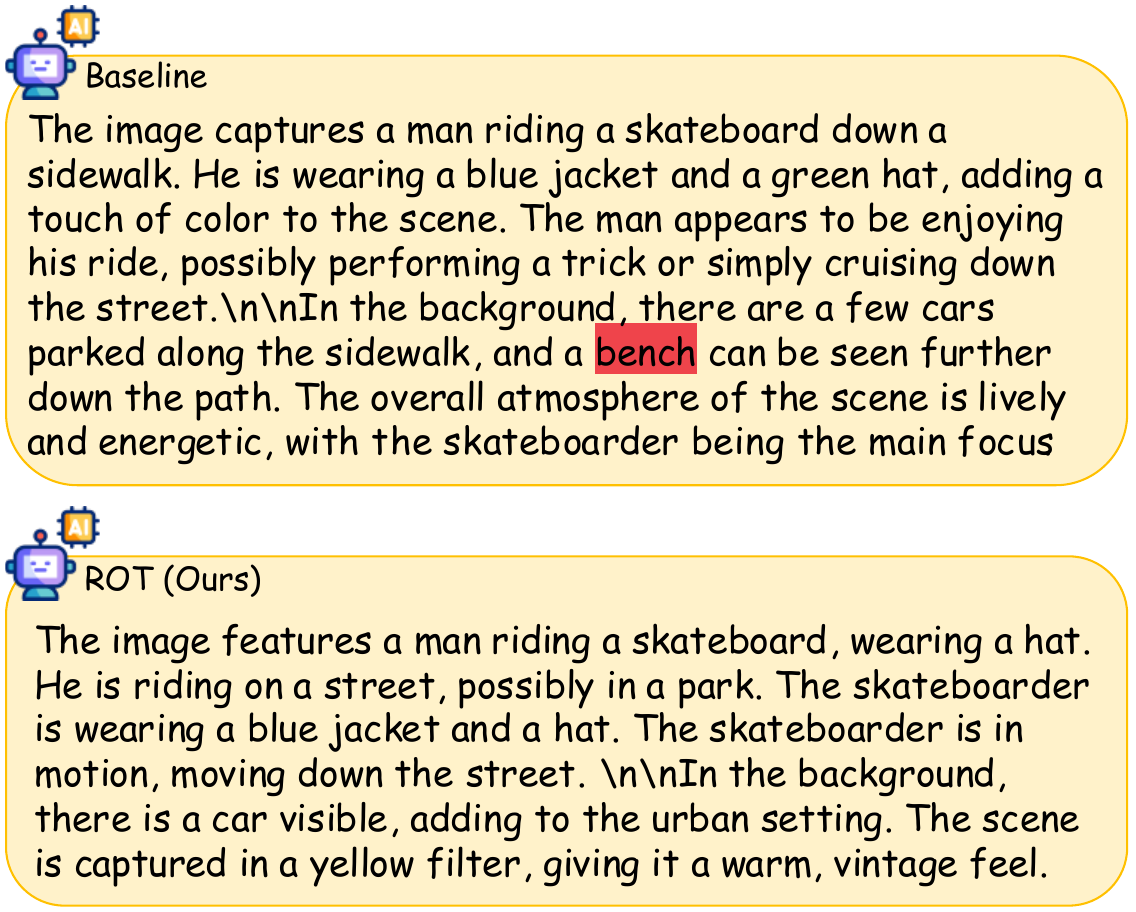}
    \end{subfigure}
    \caption{Qualitative Example 3. ROT intercepts the semantic deviation to prevent the hallucination of a ``bench''.}
    \label{fig:appendix_case3}
\end{figure}

As shown in Figure \ref{fig:appendix_case3}, the greedy strategy suffers from object hallucination by fabricating a ``bench'' in the background. ROT successfully intercepts this ungrounded semantic shift in the intermediate layers, yielding a concise and strictly factual description without introducing any non-existent instances.

\begin{figure}[H]
    \centering
    \begin{subfigure}{\columnwidth}
        \includegraphics[width=\linewidth]{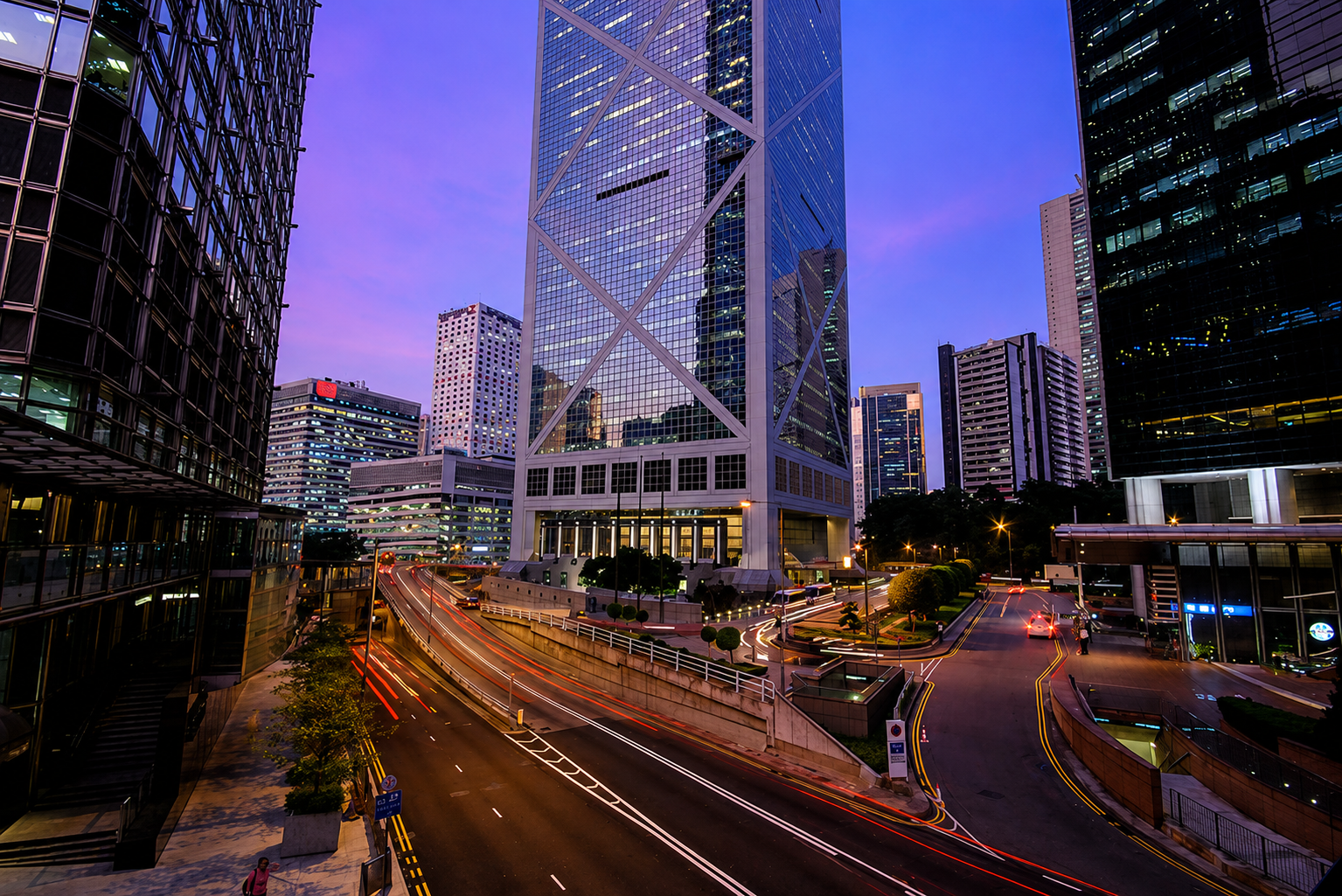}
    \end{subfigure}
    \begin{subfigure}{\columnwidth}
        \includegraphics[width=\linewidth]{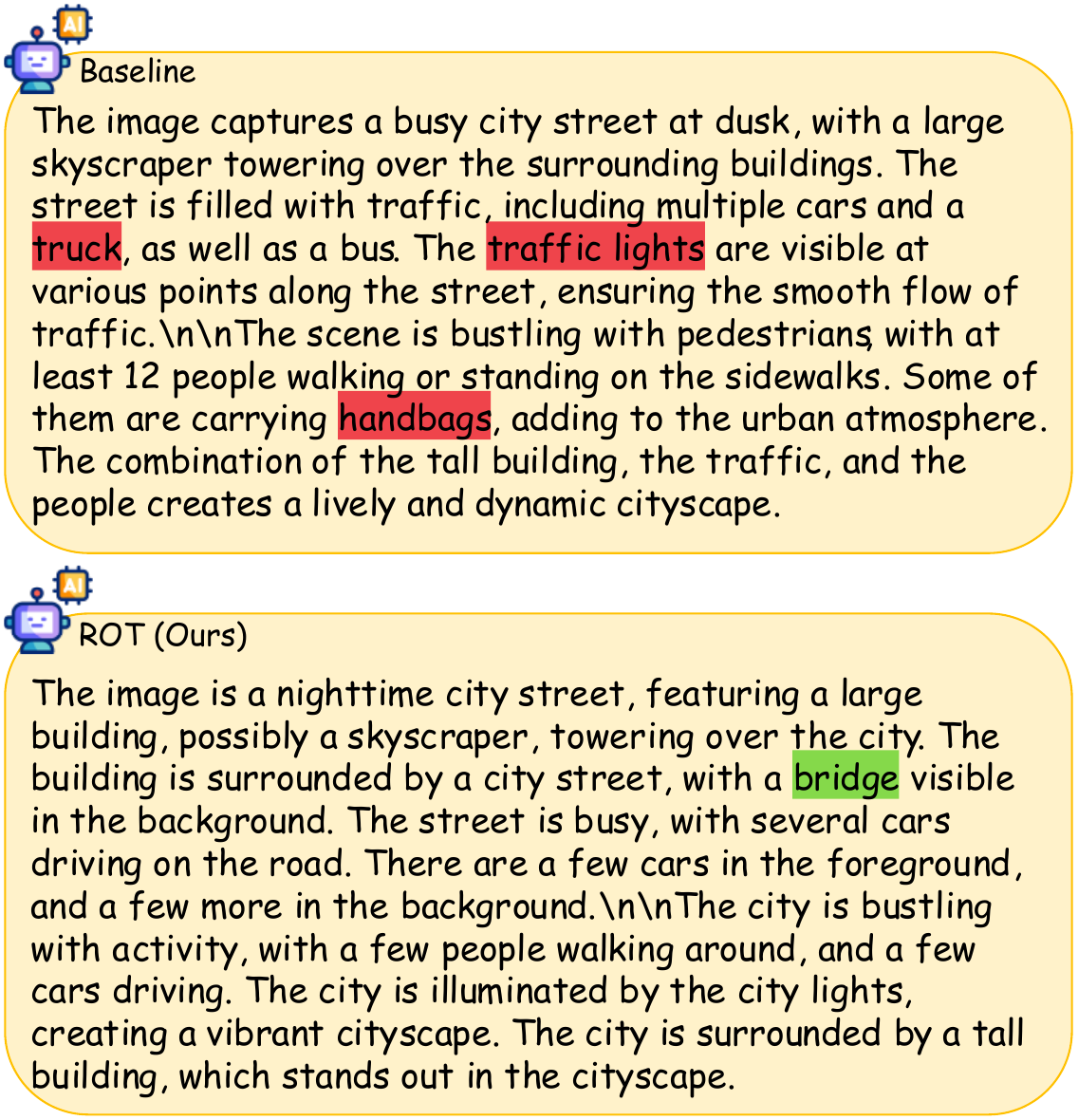}
    \end{subfigure}
    \caption{Qualitative Example 4. ROT effectively suppresses clustered hallucinations (truck, traffic lights, handbags) and correctly recognizes the ``bridge''.}
    \label{fig:appendix_case4}
\end{figure}

Figure \ref{fig:appendix_case4} demonstrates a more complex scenario where the baseline model suffers from severe multi-object hallucinations, incorrectly fabricating a ``truck'', ``traffic lights'', and ``handbags''. Conversely, ROT effectively suppresses this clustered hallucination and precisely anchors the generation to the actual visual evidence, accurately identifying the ``bridge''.

\end{document}